\documentclass[sigconf]{acmart}

\usepackage{booktabs}
\usepackage{multirow}
\usepackage{cleveref}
\usepackage{xcolor}
\usepackage{subcaption}
\usepackage{enumitem}
\usepackage{graphicx}
\usepackage{balance}

\AtBeginDocument{%
  }

\copyrightyear{2026}
\acmYear{2026}
\setcopyright{cc}
\setcctype{by}
\acmConference[CIKM '26]{Proceedings of the 35th ACM International Conference on Information and Knowledge Management}{November 07--11, 2026}{Rome, Italy}
\acmBooktitle{Proceedings of the 35th ACM International Conference on Information and Knowledge Management (CIKM '26), November 07--11, 2026, Rome, Italy}
\acmDOI{10.1145/3799682.3839968}
\acmISBN{979-8-4007-2539-5/2026/11}

\acmSubmissionID{csp3841}

\begin{document}

\title{From Counterfactual Edits to Human Narratives: LLM-Driven Interpretations for Recommender System Explanations}

\author{Vojtech Cahlik}
\authornote{Both authors contributed equally to this research.}
\correspondingauthor
\orcid{0000-0002-4222-8746}
\affiliation{%
  \institution{Faculty of Information Technology, Czech Technical University in Prague}
  \city{Prague}
  \country{Czechia}
}
\email{vojtech.cahlik@fit.cvut.cz}

\author{Amir Reza Mohammadi}
\authornotemark[1]
\orcid{0000-0003-3934-6941}
\affiliation{
    \institution{Department of Computer Science, University of Innsbruck}
    \city{Innsbruck}
    \country{Austria}
}
\email{amir.reza@uibk.ac.at}

\author{Rodrigo Alves}
\orcid{0000-0001-7458-5281}
\affiliation{%
  \institution{Faculty of Information Technology, Czech Technical University in Prague}
  \city{Prague}
  \country{Czechia}
}
\email{rodrigo.alves@fit.cvut.cz}

\author{Pavel Kordik}
\orcid{0000-0003-1433-0089}
\affiliation{%
  \institution{Faculty of Information Technology, Czech Technical University in Prague}
  \city{Prague}
  \country{Czechia}
}
\email{pavel.kordik@fit.cvut.cz}

\author{Eva Zangerle}
\orcid{0000-0003-3195-8273}
\affiliation{
    \institution{Department of Computer Science, University of Innsbruck}
    \city{Innsbruck}
    \country{Austria}
}
\email{eva.zangerle@uibk.ac.at}

\renewcommand{\shortauthors}{Vojtech Cahlik, Amir Reza Mohammadi, Rodrigo Alves, Pavel Kordik, and Eva Zangerle}

\begin{abstract}
Recommender systems increasingly influence what users discover and consume online, yet the decision logic behind their outputs often remains opaque. Counterfactual explanation (CFX) methods provide an effective way to show which changes to a user's interaction history would alter a recommendation. In recommendation settings, however, CFXs often require many edits, producing long, low-level lists that are hard for users to interpret. Moreover, most evaluations emphasize edit counts while largely overlooking whether the resulting CFXs are coherent and actionable for users. We argue that this gap between generating valid CFXs and communicating model behavior stems from the lack of user-centered abstraction and limited qualitative evaluation. To bridge this gap, we investigate large language models as mediators that summarize raw CFX interaction edit lists into concise natural language explanations describing higher-level preference patterns. We optimize these mediators for correctness, informativeness, and linguistic quality through preference optimization, and evaluate the resulting explanations with human-calibrated metrics derived from separate human-labeled datasets. On the MovieLens-1M dataset, across multiple counterfactual generators and language models, the resulting summaries achieve high overall linguistic quality, while the strongest semantic results are obtained with high-success CFX methods such as ACCENT and SPINRec. Direct Preference Optimization improves the semantic metrics in almost all evaluated settings.
\end{abstract}




\maketitle

\section{Introduction}

Recommender systems curate what users see, click, and watch~\cite{DBLP:journals/corr/abs-2201-10528,DBLP:journals/tors/BauerZS24}, yet users often have little insight into why a particular item is suggested. As recommendations become ubiquitous across online platforms, explanations that improve user trust~\cite{DBLP:conf/recsys/PeintnerMZ23}, accountability~\cite{ribeiro2016should,DBLP:journals/isci/SerARHSH24}, and informed decision-making~\cite{10.1145/3711856,10.1145/3640457.3691709} are increasingly important.

\begin{figure}[!t]
    \centering
    \captionsetup[subfigure]{skip=2pt}

    \begin{subfigure}{\columnwidth}
        \centering
        \includegraphics[width=\linewidth]{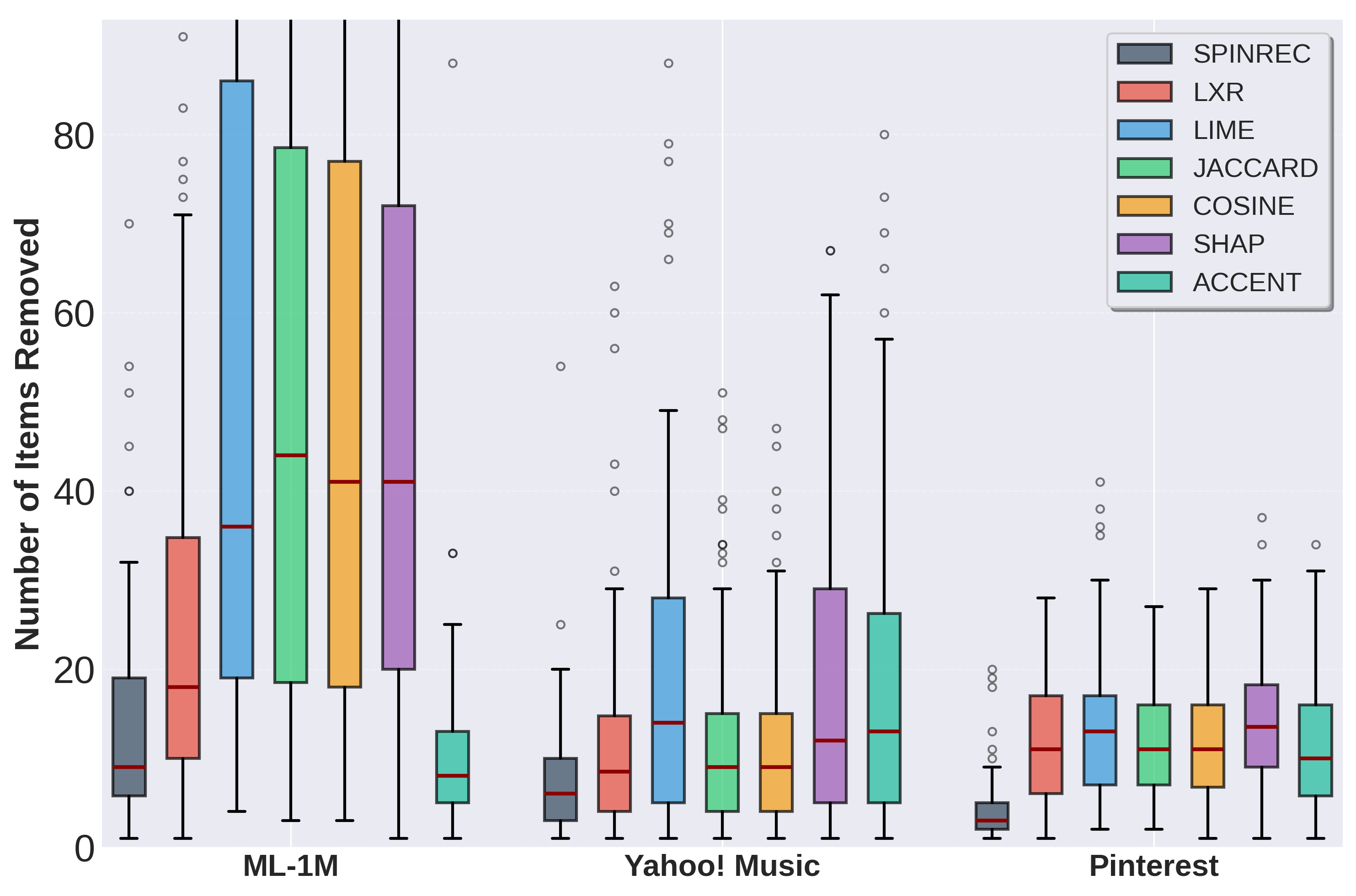}
        \caption{Number of interactions removed to displace the top-ranked item from the top-5 recommendation list.}
        \Description{Number of interactions removed to displace the top-ranked item from the top-5 recommendation list.}
        \label{fig:Compactfirst}
    \end{subfigure}

    \vspace{5pt}

    \begin{subfigure}{\columnwidth}
        \centering
        \includegraphics[width=\linewidth]{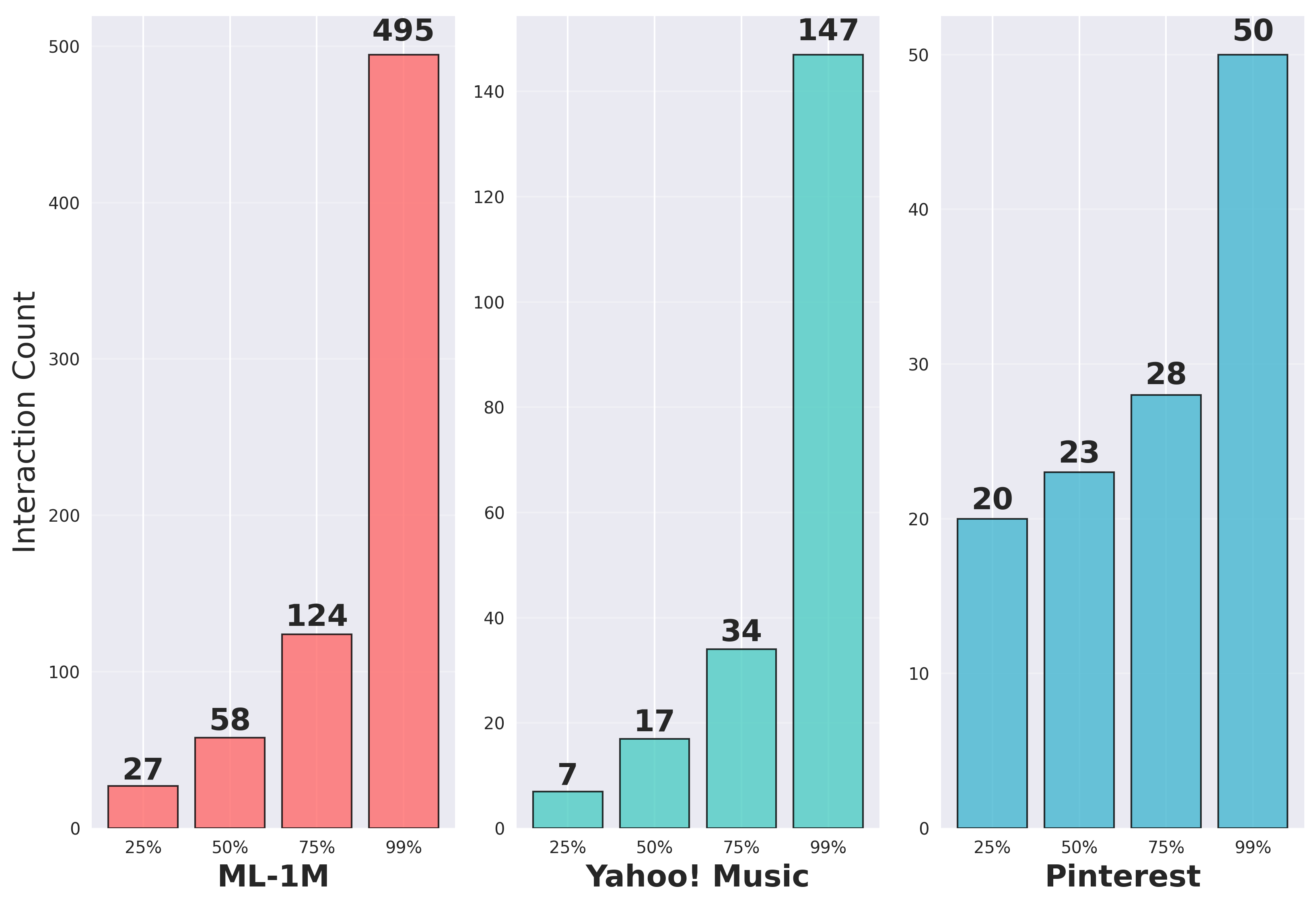}
        \caption{Percentile analysis of user history size (25\textsuperscript{th}, 50\textsuperscript{th}, 75\textsuperscript{th}, and 99\textsuperscript{th} percentiles).}
        \Description{Percentile analysis of user history size (25th, 50th, 75th, and 99th percentiles).}
        \label{fig:Compactsecond}
    \end{subfigure}

    \caption{Compactness analysis of counterfactual explanations of our variational autoencoder recommender across various datasets.
    (a) Counterfactual explanations are often composed of a large number of interactions.
    (b) These interactions often represent a substantial share of a typical user's history.}
    \Description{Compactness analysis of counterfactual explanations of a variational autoencoder recommender across various datasets.
    (a) Counterfactual explanations are often composed of a large number of interactions.
    (b) These interactions often represent a substantial share of a typical user's history.}
    \label{fig:compactnessMotivation}
\end{figure}

Counterfactual explanations (CFXs) have recently emerged as a promising mechanism for transparency in recommender systems~\cite{DBLP:journals/corr/abs-1711-00399}. By identifying minimal changes to a user's interaction history that would alter the recommendation outcome, they offer actionable ``what-if'' reasoning aligned with human intuition~\cite{DBLP:journals/mima/Freiesleben22,DBLP:conf/iccbr/AryalK24a}. However, in practice, CFXs for recommenders often require modifying a large number of historical interactions. As illustrated in~\Cref{fig:compactnessMotivation}, altering even a single top-ranked recommendation in the MovieLens-1M dataset\footnote{\url{https://grouplens.org/datasets/movielens/1m/}} (ML-1M) may require removing dozens of past interactions. This observation exposes a fundamental mismatch between \emph{algorithmic minimality} and \emph{human interpretability}~\cite{DBLP:journals/corr/abs-1711-00399}. Although a counterfactual may be minimal under an optimization objective, it is typically presented as a list of low-level perturbations such as item removals. Long edit sets can therefore overwhelm users and limit practical use. This raises the core challenge we address: how can large counterfactual edit sets be turned into concise natural language explanations (NLEs) while preserving the counterfactual grounding?

We propose an approach in which large language models (LLMs) act as mediators between counterfactual generators and users. LLMs are well suited to this role because, unlike fixed heuristics, they offer flexibility in how raw interaction data is communicated and can handle free-form text descriptions. Rather than replacing counterfactual reasoning, the mediator summarizes raw interaction edits into concise, human-centered NLEs. This mediation layer is designed to make the explanations comprehensible while preserving the underlying counterfactual grounding.

Beyond explanation generation, we also address evaluation. Existing metrics largely focus on edit counts or success rates and provide limited insight into interpretability. We use three established qualitative criteria: \textit{Correctness} and \textit{Informativeness}, which we decompose into simple interaction-level matching judgments, and \textit{Linguistic Quality}, which is scored by a prompted LLM rubric during preference optimization and by a separate human-calibrated evaluator in the final reported results.

Our contributions are (i) an LLM mediator framework that summarizes counterfactual explanations into faithful narratives, (ii) an evaluation methodology that combines decomposed interaction-level checks with human-calibrated final evaluation, and (iii) an end-to-end comparison across various CFX generators and LLMs. Experiments on the ML-1M dataset demonstrate that LLM-mediated summaries make counterfactual explanations comprehensible while preserving the counterfactual grounding. The findings suggest that combining causal counterfactual reasoning with summarization is a practical step toward more trustworthy and actionable explanations for recommender-system users. To support reproducibility, we publicly release our code and experimental pipeline\footnote{\label{foot:repo}\url{https://github.com/vcahlik/from-counterfactual-edits-to-human-narratives}}.

\section{Related Work}

Explainable AI for recommender systems has attracted growing attention due to its importance for transparency, user trust, and regulatory compliance~\cite{bilgic2005explaining,herlocker2000explaining}. Prior work falls into four broad categories: model-specific and post-hoc explanations, counterfactual explanations, side-information-based explanations, and natural language explanation methods.

\paragraph{Model-Specific and Post-Hoc Explanations}
Model-specific methods embed interpretability into the model architecture. Early examples include explainable matrix factorization models that align latent factors with interpretable components~\cite{abdollahi2016explainable,abdollahi2017using,koren2009matrix}. Recent methods such as ProtoMF~\cite{10.1145/3523227.3546756} and AI2V++~\cite{gaiger2023not} leverage prototypes or attention to highlight representative interactions. Post-hoc approaches explain predictions while treating the model as a black box: LIME~\cite{ribeiro2016should} fits local surrogate models, SHAP~\cite{DBLP:conf/nips/LundbergL17} estimates feature attributions, and influence-based methods such as FIA~\cite{cheng2019incorporating} and ACCENT~\cite{DBLP:conf/sigir/TranGR21} measure training interaction impact. Similarity heuristics (Jaccard, Cosine)~\cite{DBLP:journals/isci/BagKT19,singh2020movie} are efficient but rely on item similarity rather than on the recommender's own decision process. Overall, these methods typically output scores or long interaction lists that remain difficult to interpret.

\paragraph{Counterfactual Explanations}
CFXs identify minimal edits to a user's history that change recommendation outcomes~\cite{DBLP:conf/sigir/TranGR21,DBLP:conf/www/BarkanBGAK24}, offering causal and actionable insights compared to attribution methods. \emph{Jaccard}~\cite{DBLP:journals/isci/BagKT19} and \emph{Cosine}~\cite{singh2020movie} \emph{similarities} explain recommendations by comparing target items with a user's history via interaction-based similarity. \emph{LIME-RS}~\cite{DBLP:conf/sac/NobregaM19} locally approximates the recommender with an interpretable linear model, while \emph{SHAP}~\cite{DBLP:conf/sac/ZhongN22} assigns feature contributions based on Shapley values, albeit at high computational cost. \emph{ACCENT}~\cite{DBLP:conf/sigir/TranGR21} extends influence functions to produce actionable counterfactual explanations for neural recommenders, \emph{LXR}~\cite{DBLP:conf/www/BarkanBGAK24} uses self-supervised gradient masks to identify key interactions without perturbations, and \emph{SPINRec}~\cite{DBLP:journals/corr/abs-2511-18047} integrates gradients from realistic user baselines and selects attributions that maximize counterfactual fidelity for stable, faithful explanations. We note that the recently proposed DiceRec~\cite{DBLP:conf/sigir/MohammadiPMZ26} method was not yet available at the time this research was conducted and therefore is not included in our comparisons. Existing evaluations still focus mainly on success rates and edit counts, with much less attention to interpretability.

\paragraph{Side-Information Explanations}
Other approaches generate explanations from metadata, reviews, or knowledge graphs. Aspect-based models such as EFM~\cite{zhang2014explicit} align latent factors with explicit attributes, while path-based and graph methods justify recommendations through semantic reasoning chains~\cite{chen2020try,li2021personalized}. Although expressive, these techniques rely on rich side information and are less applicable to interaction-only settings.

\paragraph{Natural Language Explanations and LLMs}
Natural language explanations present model behavior in human-friendly text and are often evaluated by plausibility, faithfulness, or readability~\cite{jacovi2020towards}. Recent LLM-based approaches utilize prompting, synthetic rationales, or explanation supervision~\cite{huang2023can,chen2025towards,camburu2018snli,rajani2019explain,lyu2023faithful}. In the recommendation domain, CAVIAR~\cite{DBLP:conf/www/JainSS24} generates natural language explanations by describing perturbations of the recommended item's image. FIRE~\cite{sani2025fire} similarly uses attribution signals to guide LLM-generated explanations, improving linguistic quality but remaining grounded in correlational importance rather than causal interventions.

\section{Methodology}
\label{sec:method}

We connect counterfactual reasoning with natural language explanations. Unlike attribution- or similarity-based methods, we start from causally grounded counterfactual edits of user interaction history and condition the LLM directly on counterfactual interactions and samples from the user's remaining history to achieve semantic faithfulness.

Our framework consists of four stages:
\begin{enumerate}
    \item generation of counterfactual interaction-level edits using an existing CFX method;
    \item sampling of (i) counterfactual interactions and (ii) interactions from the rest of the user's history, and their conversion to item-level metadata;
    \item generation of the natural language explanation using an LLM conditioned on both sets; and
    \item evaluation and optional preference optimization of the generated explanations.
\end{enumerate}
This pipeline turns low-level counterfactual edits into concise semantic explanations while preserving their counterfactual grounding.

\subsection{Basic Notation}
Let $\mathcal{U}$ and $\mathcal{I}$ denote the sets of users and items. Each user $u \in \mathcal{U}$ is associated with an interaction history
$H_u \subseteq \mathcal{I}$.
A recommender model $f_\theta$ scores items $i \in \mathcal{I}$ for a user as $s_\theta(i \mid H_u)$, and the induced ranking defines the top-$K$ recommendation list $\mathrm{Top}_K(H_u) \subseteq \mathcal{I}$.

A counterfactual explanation $C_u$ for user $u$ is a subset of historical interactions, $C_u \subseteq H_u$, whose removal changes the recommendation outcome. In this paper, we consider interventions that displace the originally top-ranked item $i_u^\star$ from the top-$K$ recommendation list, i.e., such that $i_u^\star \notin \mathrm{Top}_K(H_u \setminus C_u)$.
We denote non-CFX interactions by $\bar C_u = H_u \setminus C_u$.

\subsection{From Counterfactual Edits to Natural Language Explanations}
\label{subsec:cf_to_nle}
Our goal is to bridge the gap between CFX outputs, which often take the form of long lists of interaction edits, and the way humans typically reason about preferences.
We introduce an LLM mediator that generates a natural language explanation
$T$ from (i) the metadata of a subset of CFX interactions and (ii) the metadata of a subset of non-CFX interactions, i.e., the user's remaining history:
\begin{equation}
T = \mathrm{LLM}_{\mathrm{gen}}\left(\phi(C_{u,k_{\mathrm{gen}}}), \phi(\bar C_{u,k_{\mathrm{gen}}})\right),
\end{equation}
where $\phi(i)$ maps each interaction item or set of items to metadata, $\mathrm{LLM}_{\mathrm{gen}}$ is a prompted LLM call that generates a natural language explanation from counterfactual and non-counterfactual interactions, and $C_{u,k_{\mathrm{gen}}}$ (resp.\ $\bar C_{u,k_{\mathrm{gen}}}$) denotes a uniform sample of $k_{\mathrm{gen}}$ interactions from $C_u$ (resp.\ $\bar C_u$), with $k_{\mathrm{gen}}$ controlling how many items are provided to the generator prompt.

\subsection{Evaluation Metrics}
\label{subsec:metrics}
\paragraph{(1) Natural Language Explanation Correctness}
We evaluate the correctness of an NLE $T$ by measuring how well it matches a sample of the user's CFX interactions:
\begin{equation}
\mathrm{Cor}(T, C_u) =
\frac{1}{|C_{u,k_{\mathrm{eval}}}|} \sum_{i \in C_{u,k_{\mathrm{eval}}}}
\mathrm{score}_{\mathrm{match}}\bigl(T, \phi(i)\bigr),
\label{eq:correctness}
\end{equation}
where $\mathrm{score}_{\mathrm{match}}(T,\phi(i))\in[0,1]$ scores how well the explanation $T$ matches item metadata $\phi(i)$, and $k_{\mathrm{eval}}$ denotes the number of sampled interactions from each set used for evaluation. During optimization, this match score is produced by a prompted LLM judge; in the final evaluation, it is produced by a human-calibrated binary classifier trained on explanation--interaction pairs. By construction, $\mathrm{Cor}(T,C_u)\in[0,1]$.

\paragraph{(2) Natural Language Explanation Informativeness}
Intuitively, an informative explanation should emphasize properties that distinguish the CFX interactions from the remaining user history. For example, an explanation that strongly matches both interaction sets is overly generic and therefore uninformative. We therefore evaluate informativeness by measuring the contrast between how well the explanation matches a sample of the CFX interactions and how well it matches a sample of the non-CFX interactions:
\begin{equation}
\mathrm{Info}(T, C_u, \bar C_u) = \mathrm{Cor}(T, C_u) - \frac{1}{|\bar C_{u,k_{\mathrm{eval}}}|} \sum_{j \in \bar C_{u,k_{\mathrm{eval}}}}
\mathrm{score}_{\mathrm{match}}\bigl(T, \phi(j)\bigr).
\label{eq:informativeness}
\end{equation}
By construction, $\mathrm{Info}(T,C_u,\bar C_u)\in[-1,1]$, where values near zero indicate generic explanations that match both sets similarly.

\paragraph{(3) Linguistic Quality of Natural Language Explanations}
We define linguistic quality as an overall score assigned directly to the explanation, $\mathrm{LQ}(T) = \mathrm{score}_{\mathrm{ling}}(T) \in [0,1]$. During DPO optimization, this signal is produced by a prompted six-criterion LLM judge rubric covering fluency, grammar, length, naturalness, illustrativeness, and specificity. In the final reported evaluation, however, we use a human-calibrated binary classifier trained on overall linguistic-quality judgments, so the reported metric is a direct calibrated binary score instead of an output of a decomposed human rubric. This keeps the human annotations holistic and authentic, rather than forcing separate criterion-level labels.


\begin{table*}[!htbp]
  \centering
  \caption{Comparison of counterfactual (CFX) generators and the corresponding LLM mediators on ML-1M,
  averaged across users (mean $\pm$ std).
  CFX Success Rate, CFX Size, and Simple CFX Rate characterize the CFX generator and are defined in \Cref{subsec:setup};
  Correctness, Informativeness, and Linguistic Quality are defined in \Cref{subsec:metrics} and are reported here
  using the human-calibrated evaluators, separately for the off-the-shelf Vanilla LLM and its DPO-optimized variant.
  Higher ($\uparrow$) or lower ($\downarrow$) values are better.
  Within each column, the best result is in \textbf{bold} and the second-best is \underline{underlined}; ties were decided using unrounded values.}
  \label{tab:main_results_new}
  \footnotesize
  \setlength{\tabcolsep}{4pt}
  
  \begin{tabular}{lccc l cc cc cc}
  \toprule
  
  \multirow{2}{*}{\textbf{\shortstack{CFX\\Generator}}} &
  \multirow{2}{*}{\textbf{\shortstack{CFX Success\\Rate $\uparrow$}}} &
  \multirow{2}{*}{\textbf{\shortstack{CFX\\Size $\downarrow$}}} &
  \multirow{2}{*}{\textbf{\shortstack{Simple CFX\\Rate $\uparrow$}}} &
  \multirow{2}{*}{\textbf{LLM}} &
  \multicolumn{2}{c}{\textbf{Correctness $\uparrow$}} &
  \multicolumn{2}{c}{\textbf{Informativeness $\uparrow$}} &
  \multicolumn{2}{c}{\textbf{Linguistic Quality $\uparrow$}} \\

  \cmidrule(lr){6-7}
  \cmidrule(lr){8-9}
  \cmidrule(lr){10-11}

  & & & & &
  \textbf{Vanilla LLM} & \textbf{DPO} &
  \textbf{Vanilla LLM} & \textbf{DPO} &
  \textbf{Vanilla LLM} & \textbf{DPO} \\
  
  \midrule
  
  \multirow{3}{*}{Jaccard}
  & \multirow{3}{*}{0.95} & \multirow{3}{*}{16.38 $\pm$ 11.67} & \multirow{3}{*}{0.02} & Ministral 8B & 0.34 $\pm$ 0.19 & 0.41 $\pm$ 0.20 & 0.14 $\pm$ 0.19 & 0.12 $\pm$ 0.20 & \underline{0.88 $\pm$ 0.33} & 0.75 $\pm$ 0.44 \\
  &  &  &  & Gemma 3 12B & 0.39 $\pm$ 0.21 & \underline{0.51 $\pm$ 0.18} & 0.13 $\pm$ 0.20 & 0.17 $\pm$ 0.23 & 0.72 $\pm$ 0.45 & 0.84 $\pm$ 0.37 \\
  &  &  &  & Qwen3 8B & 0.34 $\pm$ 0.19 & 0.46 $\pm$ 0.20 & 0.13 $\pm$ 0.19 & 0.18 $\pm$ 0.24 & 0.85 $\pm$ 0.36 & 0.71 $\pm$ 0.46 \\
  
  \midrule
  
  \multirow{3}{*}{Cosine}
  & \multirow{3}{*}{0.96} & \multirow{3}{*}{16.20 $\pm$ 11.23} & \multirow{3}{*}{0.02} & Ministral 8B & 0.34 $\pm$ 0.19 & 0.40 $\pm$ 0.21 & 0.16 $\pm$ 0.18 & 0.17 $\pm$ 0.21 & \textbf{0.91 $\pm$ 0.29} & 0.78 $\pm$ 0.41 \\
  &  &  &  & Gemma 3 12B & 0.38 $\pm$ 0.21 & 0.41 $\pm$ 0.22 & 0.13 $\pm$ 0.19 & 0.14 $\pm$ 0.21 & 0.73 $\pm$ 0.44 & 0.78 $\pm$ 0.42 \\
  &  &  &  & Qwen3 8B & 0.34 $\pm$ 0.19 & 0.40 $\pm$ 0.19 & 0.13 $\pm$ 0.22 & 0.17 $\pm$ 0.23 & 0.85 $\pm$ 0.36 & 0.78 $\pm$ 0.41 \\
  
  \midrule
  
  \multirow{3}{*}{LIME-RS}
  & \multirow{3}{*}{0.84} & \multirow{3}{*}{18.52 $\pm$ 11.84} & \multirow{3}{*}{0.01} & Ministral 8B & 0.32 $\pm$ 0.17 & \textbf{0.52 $\pm$ 0.16} & 0.11 $\pm$ 0.19 & 0.13 $\pm$ 0.21 & 0.86 $\pm$ 0.34 & 0.59 $\pm$ 0.49 \\
  &  &  &  & Gemma 3 12B & 0.37 $\pm$ 0.21 & 0.39 $\pm$ 0.18 & 0.11 $\pm$ 0.21 & 0.14 $\pm$ 0.20 & 0.62 $\pm$ 0.49 & 0.78 $\pm$ 0.42 \\
  &  &  &  & Qwen3 8B & 0.36 $\pm$ 0.16 & 0.44 $\pm$ 0.18 & 0.10 $\pm$ 0.20 & 0.14 $\pm$ 0.22 & 0.80 $\pm$ 0.40 & 0.64 $\pm$ 0.48 \\
  
  \midrule
  
  \multirow{3}{*}{SHAP}
  & \multirow{3}{*}{0.76} & \multirow{3}{*}{20.83 $\pm$ 11.82} & \multirow{3}{*}{0.01} & Ministral 8B & 0.29 $\pm$ 0.17 & 0.32 $\pm$ 0.17 & 0.09 $\pm$ 0.17 & 0.09 $\pm$ 0.18 & 0.87 $\pm$ 0.34 & \textbf{0.86 $\pm$ 0.34} \\
  &  &  &  & Gemma 3 12B & 0.32 $\pm$ 0.20 & 0.33 $\pm$ 0.20 & 0.07 $\pm$ 0.19 & 0.08 $\pm$ 0.20 & 0.71 $\pm$ 0.46 & 0.76 $\pm$ 0.43 \\
  &  &  &  & Qwen3 8B & 0.32 $\pm$ 0.16 & 0.34 $\pm$ 0.16 & 0.09 $\pm$ 0.20 & 0.10 $\pm$ 0.20 & 0.77 $\pm$ 0.42 & 0.77 $\pm$ 0.42 \\
  
  \midrule
  
  \multirow{3}{*}{ACCENT}
  & \multirow{3}{*}{\underline{0.99}} & \multirow{3}{*}{\textbf{10.36 $\pm$ 7.24}} & \multirow{3}{*}{\textbf{0.07}} & Ministral 8B & 0.38 $\pm$ 0.19 & 0.40 $\pm$ 0.20 & \underline{0.18 $\pm$ 0.20} & 0.18 $\pm$ 0.20 & 0.84 $\pm$ 0.36 & 0.83 $\pm$ 0.37 \\
  &  &  &  & Gemma 3 12B & \textbf{0.43 $\pm$ 0.22} & 0.49 $\pm$ 0.21 & 0.16 $\pm$ 0.21 & \underline{0.21 $\pm$ 0.22} & 0.72 $\pm$ 0.45 & 0.72 $\pm$ 0.45 \\
  &  &  &  & Qwen3 8B & \underline{0.41 $\pm$ 0.19} & 0.44 $\pm$ 0.23 & \textbf{0.18 $\pm$ 0.22} & 0.21 $\pm$ 0.24 & 0.73 $\pm$ 0.44 & 0.64 $\pm$ 0.48 \\
  
  \midrule
  
  \multirow{3}{*}{LXR}
  & \multirow{3}{*}{0.93} & \multirow{3}{*}{15.91 $\pm$ 11.03} & \multirow{3}{*}{0.00} & Ministral 8B & 0.33 $\pm$ 0.18 & 0.42 $\pm$ 0.18 & 0.13 $\pm$ 0.20 & 0.19 $\pm$ 0.20 & 0.84 $\pm$ 0.36 & 0.79 $\pm$ 0.41 \\
  &  &  &  & Gemma 3 12B & 0.40 $\pm$ 0.21 & 0.48 $\pm$ 0.20 & 0.12 $\pm$ 0.20 & 0.13 $\pm$ 0.22 & 0.66 $\pm$ 0.47 & 0.84 $\pm$ 0.36 \\
  &  &  &  & Qwen3 8B & 0.36 $\pm$ 0.19 & 0.36 $\pm$ 0.19 & 0.12 $\pm$ 0.21 & 0.12 $\pm$ 0.21 & 0.84 $\pm$ 0.37 & \underline{0.85 $\pm$ 0.36} \\
  
  \midrule
  
  \multirow{3}{*}{SPINRec}
  & \multirow{3}{*}{\textbf{1.00}} & \multirow{3}{*}{\underline{12.25 $\pm$ 8.75}} & \multirow{3}{*}{\underline{0.05}} & Ministral 8B & 0.36 $\pm$ 0.19 & 0.43 $\pm$ 0.20 & 0.15 $\pm$ 0.19 & 0.15 $\pm$ 0.21 & 0.85 $\pm$ 0.36 & 0.71 $\pm$ 0.46 \\
  &  &  &  & Gemma 3 12B & 0.41 $\pm$ 0.21 & 0.50 $\pm$ 0.23 & 0.15 $\pm$ 0.20 & 0.17 $\pm$ 0.22 & 0.71 $\pm$ 0.46 & 0.65 $\pm$ 0.48 \\
  &  &  &  & Qwen3 8B & 0.39 $\pm$ 0.19 & 0.50 $\pm$ 0.21 & 0.15 $\pm$ 0.21 & \textbf{0.22 $\pm$ 0.21} & 0.75 $\pm$ 0.44 & 0.64 $\pm$ 0.48 \\
  
  \bottomrule
  \end{tabular}
\end{table*} 

\begin{table*}[!htbp]
  \centering
  \caption{Examples of natural language explanations obtained for ACCENT and DPO-tuned Ministral~8B,
  together with per-explanation evaluation scores.
  For each metric, \textbf{LLM judge} is the score obtained using the prompted LLM judge used during DPO optimization,
  and \textbf{Human-cal.} is the score obtained using the corresponding human-calibrated classifier used for the final evaluation.
  Higher values are better ($\uparrow$).}
  \label{tab:example_readability}
  \footnotesize
  \setlength{\tabcolsep}{3pt}
  
  \begin{tabular}{@{}p{7cm}cccccc@{}}
  \toprule
  \multirow{2}{*}{\textbf{Explanation}} &
  \multicolumn{2}{c}{\textbf{Correctness $\uparrow$}} &
  \multicolumn{2}{c}{\textbf{Informativeness $\uparrow$}} &
  \multicolumn{2}{c}{\textbf{Linguistic Quality $\uparrow$}} \\
  \cmidrule(lr){2-3}
  \cmidrule(lr){4-5}
  \cmidrule(lr){6-7}
  & \textbf{LLM judge} & \textbf{Human-cal.} &
  \textbf{LLM judge} & \textbf{Human-cal.} &
  \textbf{LLM judge} & \textbf{Human-cal.} \\
  \midrule
  \textit{(Because you watched)} comedy and drama films from 1990s & 0.35 & 0.53 & 0.02 & 0.08 & 0.83 & 1.00 \\
  
  \textit{(Because you watched)} adventure romances from 1995 & 0.21 & 0.12 & 0.11 & 0.12 & 0.94 & 1.00 \\
  
  \textit{(Because you watched)} drama films from 1960s to 1990s & 0.83 & 0.50 & 0.44 & 0.40 & 0.89 & 0.00 \\
  
  \textit{(Because you watched)} mystery and action films from 1970s to 1990s & 0.60 & 0.40 & 0.28 & 0.25 & 1.00 & 1.00 \\
  
  \textit{(Because you watched)} comedy films from 1980s and 1990s & 0.46 & 0.25 & -0.15 & -0.20 & 0.89 & 1.00 \\
  
  \bottomrule
  \end{tabular}
\end{table*}

\subsection{Optimization}
To improve the overall quality of the generated natural language explanations, we optimize the generation LLM using Direct Preference Optimization (DPO)~\cite{rafailov2023direct}. We construct pairwise NLE data on held-out users and score candidate explanations with prompted LLM judges using the correctness, informativeness, and linguistic-quality signals defined above. We manually refine these judge prompts on a human-labeled validation dataset to better align the optimization signal with human judgments. We then tune the main DPO hyperparameters, that is, the learning rate and the regularization parameter $\beta$, by random search under a fixed compute budget and select the final DPO variant using separate human-calibrated evaluators on held-out validation users.

\section{Experiments and Results}
\label{sec:experiments}

\subsection{Experimental Setup}
\label{subsec:setup}

We evaluate our approach on the movie-database ML-1M dataset, which represents a dense interaction setting typical of mature recommendation platforms. We train a variational autoencoder recommender~\cite{liang2018variational} on a separate training set of users and use it to generate CFXs that push the originally top-ranked item out of the top-5 recommendation list. To prevent the explanation LLM from relying on its internal knowledge when generating the NLEs, we omit item titles and use only the release year and list of genres as item features. We refer to the number of removed interactions $|C_u|$ as the \emph{CFX Size}. Cases with $|C_u|>50$ are treated as failures, as such large edit sets are unlikely to yield usable human-facing explanations; we report the fraction of successful cases as \emph{CFX Success Rate}. Additionally, we exclude CFXs with $|C_u|<3$, whose fraction with respect to all successful cases we report as the \emph{Simple CFX Rate}, since these short edit sets can usually be communicated directly from the items themselves without additional summarization. In LLM prompts, we use interaction set sample sizes of $k_{\mathrm{gen}}=k_{\mathrm{eval}}=20$.

We use Ministral 8B, Gemma 3 12B, and Qwen3 8B LLMs for NLE generation. For DPO, we construct pairwise training datasets on a validation set of users using alternative explanations generated by two LLMs. Candidate explanations are then scored by prompted LLM judges, and the main DPO hyperparameters, learning rate and $\beta$, are tuned by random search under a fixed compute budget. The DPO variant reported in the main table is selected by human-calibrated validation performance on held-out validation users.

To avoid reporting metrics used during optimization as final evaluation, the main table uses separate human-calibrated evaluators for final testing. We train ModernBERT-large binary classifiers on separate human-labeled datasets, each with 100 held-out test instances, obtaining accuracies of 0.91 for linguistic quality and 0.92 for explanation--interaction matching. Correctness and informativeness then use the same formulas as in \Cref{subsec:metrics} with classifier-based interaction scores, while linguistic quality is obtained using individual binary scores.

\subsection{Results}
\label{sec:results}

Under the human-calibrated evaluation approach, \Cref{tab:main_results_new} summarizes all results, and \Cref{tab:example_readability} shows representative ACCENT explanations from the DPO-tuned Ministral~8B model.

\paragraph{Characteristics of the CFX Generators}
ACCENT and SPINRec achieve the highest CFX success rates ($0.99$ and $1.00$, respectively) and also produce the most compact CFXs, with average sizes of $10.36$ and $12.25$ interactions. They additionally have the highest simple-CFX rates, although this rate remains low for all methods. Most cases therefore still involve larger edit sets and benefit from natural language summarization.

\paragraph{Correctness and Informativeness}
Semantic quality is highest for the stronger CFX methods, especially ACCENT and SPINRec, even though the individual best rows are spread across several method--LLM combinations. DPO improves informativeness for most method--LLM pairs, and improves correctness in all but one case, indicating that preference optimization usually helps the LLM produce summaries that are closer to the counterfactual edits and better highlight patterns that distinguish them from the rest of the user's history.

\paragraph{Linguistic Quality}
Linguistic quality remains high overall, with many settings above $0.80$ under the human-calibrated evaluator, and above $0.85$ for the strongest results. Some spread remains in these scores, which is partially expected given the multiple optimization objectives. This highlights why a careful design of the optimized metrics is important, as it must allow the optimization to improve semantic alignment without losing pressure towards readable outputs.

\section{Conclusion}

In this work, we presented a natural language explanation framework for recommender systems that summarizes raw counterfactual edits using LLMs. These explanations remain grounded in the underlying counterfactuals in terms of human-calibrated correctness and informativeness while maintaining high linguistic quality. Performance depends on both the underlying CFX generator and the language model; Direct Preference Optimization improved correctness and informativeness in nearly all of the tested settings. Overall, the results suggest that LLM-based summarization can make counterfactual explanations more accessible without losing their causal grounding.

\begin{acks}
This work was supported by the project "PoliRuralPlus: Fostering Sustainable Development of Rural-Urban Communities Using Spatial Enhanced Attractiveness Mapping ToolBox," funded by the European Union (REA), grant No. 101136910; by the FOCAL project, funded by the European Union under grant agreement No. 101137787; by the National Centre for Artificial Intelligence (NCUI), project No. TQ28000003, co-funded by the Technology Agency of the Czech Republic under the SIGMA Programme; and by the scholarship granted by the University of Innsbruck, Vice-Rectorate for Research (Doktoratsstipendium 2026). Access to the computational infrastructure of the OP VVV funded project CZ.02.1.01/0.0/0.0/16\_019/0000765 ``Research Center for Informatics'' is gratefully acknowledged. Computational resources were also provided by the e-INFRA CZ project (ID:90254), supported by the Ministry of Education, Youth and Sports of the Czech Republic.
\end{acks}

\section*{GenAI Usage Disclosure}

We used Cursor with Composer-2-series, GPT-5-series, and Claude Opus 4-series models for coding assistance and debugging. All AI-generated code suggestions were manually reviewed, and revised as necessary, before use. All experiments and analyses were conducted manually without the use of generative AI.

\bibliographystyle{ACM-Reference-Format}
\balance
\bibliography{literature}

\end{document}